\pdfoutput=1

\documentclass[preprint,11pt,authoryear]{elsarticle}

\usepackage{amssymb}
\usepackage{amsmath}
\usepackage{booktabs}
\usepackage{lmodern}
\usepackage{graphicx}
\usepackage{xcolor}
\usepackage{colortbl}
\usepackage{rotating}
\usepackage{makecell}
\usepackage{multirow}
\usepackage{amsmath}
\usepackage{hyperref}
\usepackage{natbib}
\usepackage{float}
\usepackage{pdflscape}

\journal{Science of Remote Sensing}
\makeatletter
\def\ps@pprintTitle{\let\@oddhead\@empty\let\@evenhead\@empty\let\@oddfoot\@empty\let\@evenfoot\@empty}
\makeatother

\defcitealias{EU_2024}{EU,~2024}
\defcitealias{adv_2020}{AdV,~2020}
\defcitealias{LGB_luftbilder_2024}{LGB,~2025b}
\defcitealias{LGB_nDOM_2024}{a}
\defcitealias{SH_orthophotos_2024}{LVermGeo SH,~2025b}
\defcitealias{SH_bDOM_2024}{a}
\defcitealias{Geobasis_NRW_luftbilder_2024}{Geobasis NRW,~2025a}
\defcitealias{Geobasis_NRW_nDOM_2024}{b}
\defcitealias{ISPRSUrbanSemLab}{ISPRS,~2025}

\begin{document}

\newcommand{\citealiastext}[1]{\citetalias{#1}}
\newcommand{\journalDOI}{10.1016/j.srs.2026.100510}

\begin{frontmatter}

\title{Mapping and Classification of Trees Outside Forests using Deep Learning\tnoteref{am}}
\tnotetext[am]{This is the accepted manuscript of an article published in \textit{Science of Remote Sensing} 14 (2026) 100510. The final version is available at \href{https://doi.org/\journalDOI}{https://doi.org/\journalDOI}. \copyright~2026 The Author(s). This manuscript version is made available under the CC BY 4.0 license \href{https://creativecommons.org/licenses/by/4.0/}{https://creativecommons.org/licenses/by/4.0/}.}

\author[JointLab,Osna,ATB]{Moritz Lucas}
\author[JointLab]{Hamid Ebrahimy}
\author[JointLab]{Viacheslav Barkov}
\author[ATB,JointLab]{Ralf Pecenka}
\author[CogSi,JointLab]{Kai-Uwe Kühnberger}
\author[Osna,JointLab]{Björn Waske}

\affiliation[JointLab]{organization={Osnabrück University, Joint Lab Artificial Intelligence and Data Science},
            city={Osnabrück},
            postcode={49074},
            state={Lower Saxony},
            country={Germany}}
\affiliation[ATB]{organization={Leibniz Institute for Agricultural Engineering and Bioeconomy, System Process Engineering},
            postcode={14469},
            state={Brandenburg},
            country={Germany}}
\affiliation[CogSi]{organization={Osnabrück University, Institute of Cognitive Science, Artificial Intelligence},
            postcode={49090},
            state={Lower Saxony},
            country={Germany}}
\affiliation[Osna]{organization={Osnabrück University, Institute for Computer Science, Remote Sensing Osnabrück},
            city={Osnabrück},
            postcode={49074},
            state={Lower Saxony},
            country={Germany}}

\begin{abstract}
Trees Outside Forests (TOF) play an important role in agricultural landscapes by supporting biodiversity, sequestering carbon, and regulating microclimates. Yet, most studies have treated TOF as a single class or relied on rigid rule-based thresholds, limiting ecological interpretation and adaptability across regions. To address this, we evaluate deep learning for TOF classification using a newly generated dataset and high-resolution aerial imagery from four agricultural landscapes in Germany. Specifically, we compare convolutional neural networks (CNNs), vision transformers, and hybrid CNN–transformer models across six semantic segmentation architectures (ABCNet, LSKNet, FT-UNetFormer, DC-Swin, BANet, and U-Net) to map four structural classes: Forest, Patch, Linear, and Tree, derived from previous studies and governmental products.
Overall, the models achieved good classification accuracy across the four landscapes, with the FT-UNetFormer performing best (mean Intersection-over-Union 0.74; mean F1 score 0.84), underscoring the importance of spatial context understanding in TOF mapping and classification. Our results show good results for Forest and Linear class and reveal challenges particularly in classifying complex structures with high edge density, notably the Patch and Tree class. Our generalization experiments highlight the need for regionally diverse training data to ensure reliable large-scale mapping. The dataset and code are openly available at \href{https://github.com/Moerizzy/TOFMapper}{https://github.com/Moerizzy/TOFMapper}.

\end{abstract}


\end{frontmatter}

\section{Introduction}

Trees Outside Forests (TOF) comprise all trees growing beyond the forest definition of the Food and Agriculture Organization (FAO) \citep{FAO_2013}, and they take a range of structural forms, from hedgerows and small groves to individual farmland trees. TOF are an integral component of urban and agricultural landscapes and provide essential ecosystem services \citep{Peros_2022}. They support biodiversity by offering habitat and improving connectivity in fragmented landscapes \citep{Prevedello_2018}. In agricultural areas, they enhance soil quality through carbon sequestration and reduce soil erosion by lowering wind speeds and surface water runoff \citep{Mayer_2022, Bohm_2014, Fahad_2022}. They also moderate the local microclimate by increasing soil moisture retention and water-holding capacity and by buffering evapotranspiration and temperature extremes \citep{Kanzler_2019, Rivest_2022, Aalto_2022}. Together, these functions increase the climate resilience of agricultural landscapes and reduce crop vulnerability to extreme weather \citep{Ivezic_2021, MosqueraLosada_2018}. In countries with low forest cover and highly fragmented forests, TOF add substantially to national carbon stocks \citep{Liu_2023}. In historically evolved agricultural landscapes, they also hold aesthetic and cultural value as elements of traditional land use and contribute to recreational quality \citep{Oreszczyn_2000, Kienast_2012}.
These functions depend on the structural type of TOF. Linear hedgerows are particularly important for connectivity and as windbreaks, larger groves for as wildlife refuge, and individual trees for local microclimate and landscape character. Assessing TOF therefore requires data that not only detects tree cover outside forests but also distinguishes its structural forms. Such detailed and spatially comprehensive data remain limited, yet they are critical for ecological monitoring, conservation, and informed land-use planning \citep{Peros_2022}.

Traditionally, the assessment of TOF relied on field-based inventories. These are accurate but costly, labor-intensive, and limited in spatial coverage, and they frequently underestimate the true abundance and distribution of TOF \citep{Schnell_2015}. Remote sensing offers a scalable alternative that enables continuous and spatially comprehensive mapping \citep{lodato_2025}. Early studies classified TOF using height and spectral thresholds \citep{Meneguzzo_2013, Pujar_2014}. \cite{Bolyn_2019} classified TOF into six structural classes from aerial imagery over a 20 km² area, and \cite{Malkoc_2021} mapped TOF across Switzerland from land cover and height thresholds, among the first large-scale efforts enabled by high-resolution height models \citep{Maack_2017}. A common limitation of these approaches is that the heuristic thresholds are tuned to a specific landscape and do not transfer well to others \citep{Bolyn_2019}.
Deep learning subsequently improved the scalability and transferability of TOF mapping, owing to its ability to learn features that adapt across environmental contexts \citep{Kattenborn_2021}. The fusion of Sentinel-1 and Sentinel-2 data enabled the first global assessment of TOF \citep{Brandt_2021}, and high-resolution satellite imagery enabled detailed mapping across India, Africa, and Europe \citep{Brandt_2024, Reiner_2023, Liu_2023}. To keep annotation tractable and to maximize generalization, however, these studies reduce TOF to a binary distinction between trees inside and outside forests. This gain in scalability comes at the cost of structural detail, as the resulting maps cannot separate hedgerows, groves, and individual farmland trees, which limits their value for habitat modeling and for auditing agricultural and environmental policy \citep{Dronova_2022}.
To recover this detail, recent studies target individual TOF structures \citep{Muro_2025, Li_2023_tree}. \cite{Huber-Garcia_2025} mapped hedgerows in Bavaria with a DeepLabV3 segmentation model and aerial imagery, and showed that the Copernicus Land Monitoring Service High Resolution Layer Small Woody Features (SWF) \citep{CopernicusSWF_2024} misses finer-scale features that high-resolution aerial imagery resolves. Such studies demonstrate the benefit of high-resolution imagery, but each addresses only a single structure type and does not provide a unified classification of the different TOF structures.
A gap therefore remains between scalable but coarse binary maps and detailed maps of single structures. To our knowledge, no study has systematically mapped the full range of TOF structural classes in a way that transfers across landscapes. At the same time, semantic segmentation architectures that model spatial context, in particular vision transformers and modern CNNs, achieve strong results on remote sensing benchmarks (\citealp{Wang_2023}; \citealiastext{ISPRSUrbanSemLab}) and are well suited to learning the complex spatial relationships that distinguish TOF structures \citep{Zang_2021}. The growing availability of high-resolution aerial imagery, promoted by initiatives such as the EU INSPIRE directive \citepalias{EU_2024}, makes their large-scale application increasingly feasible. These architectures have not yet been systematically evaluated for multiclass TOF mapping.
This study addresses that gap. We evaluate state-of-the-art CNN and vision transformer architectures for the semantic segmentation and classification of TOF from high-resolution RGB aerial imagery across four distinct agricultural landscapes in Germany. Following established definitions, we assign tree cover to four geometric classes: \textit{Forest}, \textit{Patch}, \textit{Linear}, and \textit{Tree} \citep{FAO_2013, Bolyn_2019, MaurizioSarti_2021, CopernicusSWF_2024}. Our approach combines the structural detail of single-feature studies with the transferability of deep learning, which earlier work achieved only separately. Specifically, we (i) provide the first systematic evaluation of modern segmentation architectures for multiclass TOF mapping, (ii) assess their spatial transferability to regions not seen during training, and (iii) show that small TOF elements, including individual trees and narrow hedgerows, can be mapped reliably from RGB imagery alone, enabling fine-grained ecological analysis.

Our contributions include: (1) presenting the first application of deep learning for TOF classification, (2) comparing CNN-, transformer-, and hybrid-based semantic segmentation architectures to identify the most effective model for TOF mapping and classification, (3) analyzing accuracy across TOF classes and landscape structures to highlight strengths and weaknesses, and (4) assessing the spatial generalization of the best-performing model to evaluate its suitability for large-scale applications within Germany.

\newpage

\section{Study Area and Datasets}

\subsection{Study Areas}

\begin{figure}[ht]
    \centering
    \includegraphics[width=1\textwidth]{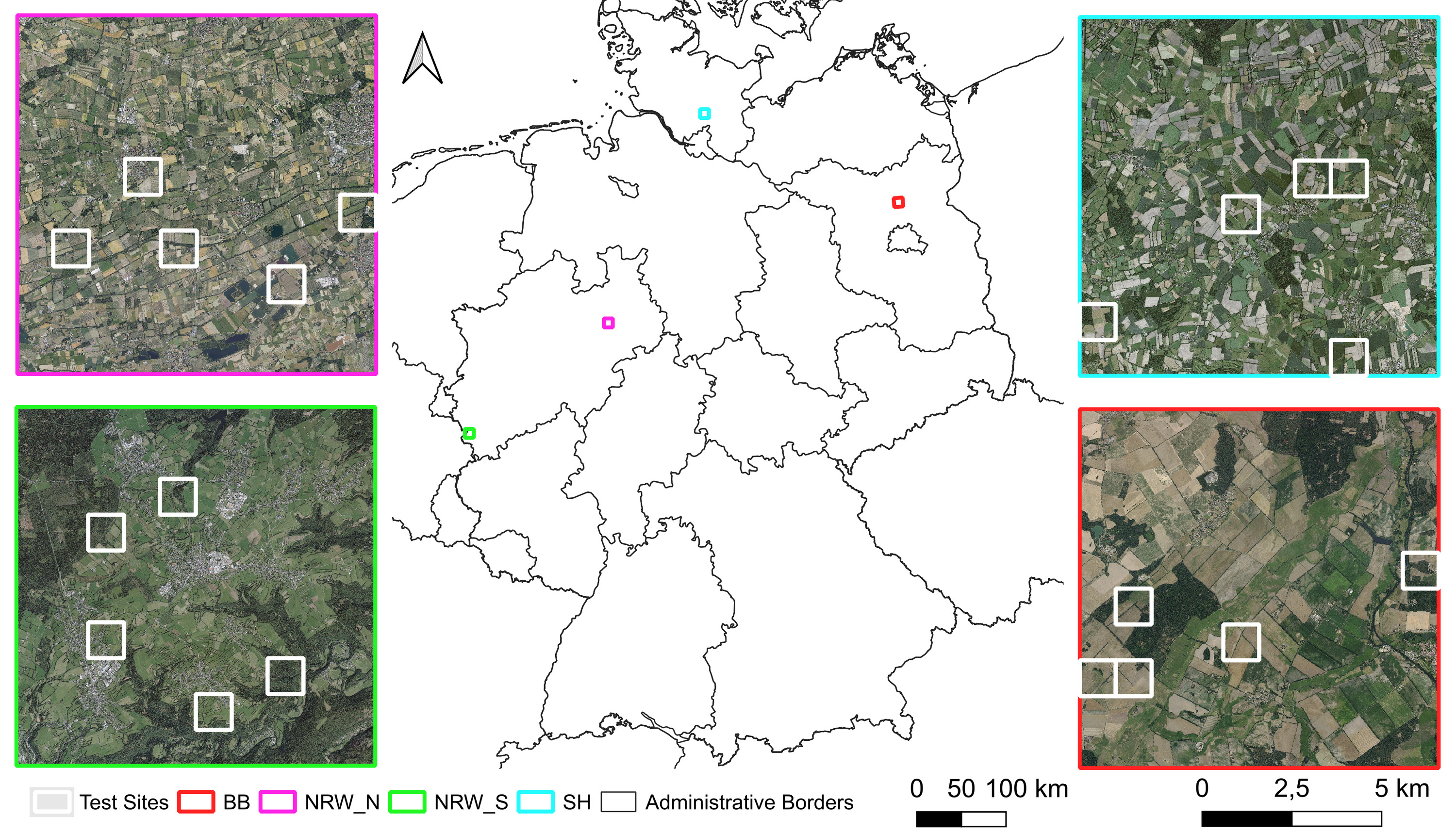}
    \caption{Location of the study areas in Germany with detailed insets for each region. The four study areas—Brandenburg (BB), North Rhine-Westphalia North (NRW\_N), North Rhine-Westphalia South (NRW\_S), and Schleswig-Holstein (SH) are outlined in different colors. White boxes indicate the locations of the representative testing tiles.}
    \label{fig:study_areas}
\end{figure}

In order to test the methodology in different environments and to assess its spatial generalization, four study areas with distinct landscape structures were selected (Figure~\ref{fig:study_areas}~\&~Table~\ref{tab:dataset_metadata}). The selected areas capture the range of agricultural landscapes present in Germany. In northern Germany, the Schleswig-Holstein (SH) study area is distinguished by low tree cover and a high proportion of TOF. This region is characterized by hedgerow structure and traditional smallholder farming practices \citep{Meynen_1962, Grajetzky_1993}. In the east, the Brandenburg (BB) study area is dominated by very large agricultural plots shaped by collectivized agriculture, resulting in the lowest TOF rate among the study areas (Table~\ref{tab:dataset_metadata}) \citep{Wolff_2021}. In central Germany, the North Rhine-Westphalia North (NRW\_N) study area represents a smallholder agricultural landscape with intensive farming. In contrast, North Rhine-Westphalia South (NRW\_S) is a more hilly region with elevation ranging from 300 to 500 m above sea-level, characterized by extensive grasslands and forests. All study areas share the same dimensions, each forming a square with a side length of 10 km, resulting in an area of 100 km² per study site.

\begin{table}[ht]
    \centering
    \caption{Metadata of the study areas used for training, validation, and testing, including acquisition date, overall tree cover, and the distribution into \textit{Forest} and TOF into \textit{Patch}, \textit{Linear}, and \textit{Tree} classes.}
    \begin{tabular}{lllllllll}
        \toprule
        \multirow{2}{*}{State} & Acquisition & Tree  & Forest & TOF  & Patch & Linear & Tree \\
                               & Date        & Cover(\%)       & (\%)   & (\%) & (\%)  & (\%)   & (\%) \\
        \midrule
        BB                     & 19.07.2022  & 19.4       & 16.5   & 2.9  & 0.6   & 1.7    & 0.6  \\
        NRW\_N                 & 15.06.2022  & 13.1       & 7.4    & 5.7  & 1.5   & 2.9    & 1.3  \\
        NRW\_S                 & 11.08.2023  & 44.0       & 38.2   & 5.8  & 1.2   & 3.1    & 1.5  \\
        SH                     & 14.06.2021  & 14.2       & 7.7    & 6.5  & 1.2   & 3.9    & 1.4  \\
        \midrule
        Overall                    & --          & 22.7 & 17.5 & 5.2 & 1.1 & 2.9 & 1.2 \\
        \bottomrule
    \end{tabular}
    \label{tab:dataset_metadata}
\end{table}

\subsection{Aerial Imagery and Height Models}

The database consists of Digital Orthophotos (DOP) and photogrammetrically derived normalized Digital Surface Models (nDSM), both captured on the same dates. All datasets are freely available, with those for BB and NRW distributed under the Data Licence Germany (dl-de/by-2-0) and those for SH under Creative Commons (CC BY 4.0). The datasets were acquired by the respective federal states agencies~\citepalias{LGB_luftbilder_2024, LGB_nDOM_2024, SH_orthophotos_2024, SH_bDOM_2024, Geobasis_NRW_luftbilder_2024, Geobasis_NRW_nDOM_2024}.
All DOP were collected during the summer months (June - August) between 2021 and 2023 and include four spectral bands (RGB and near-infrared). The RGB information was used for deep learning based TOF mapping and classification, while the additional near-infrared band and nDSM were utilized for creating the reference data. Government-acquired DOP must meet quality standards \citepalias{adv_2020}, ensuring distortion-free imagery, positional inaccuracies of less than 0.4~m, and an 8-bit color depth, corresponding to a scale between 0 and 255 including continuous relative radiometric correction. These standards enable precise, positionally accurate image analysis with consistent quality across datasets \citepalias{adv_2020}.
To ensure comparability, the raster data were resampled to a spatial resolution of 20 cm using nearest-neighbor interpolation, preserving the original pixel values during the resampling process.

\newpage

\section{Methodology}

\subsection{Generation of Reference Data}
\label{sec:generation_reference_data}

We define four classes based on the SWF and established TOF typologies  (\citealp{Straub_2008, Pujar_2014, Bolyn_2019, MaurizioSarti_2021, CopernicusSWF_2024}). Unlike the SWF dataset, which does not resolve individual trees at its spatial resolution, our scheme includes a dedicated Tree class. The defining thresholds follow these studies and were refined on our data.
Based on these considerations, the polygons were categorized into the following classes:
\begin{itemize}
    \item \textit{Forest}: Wooded areas that are larger than 5,000~m² and elongation ratio (length/width) smaller than 3.
    \item \textit{Patch}: Areas between 500~m² and 0.5~ha with an elongation ratio smaller than 3.
    \item \textit{Linear}: Areas with an elongation ratio greater than 3.
    \item \textit{Tree}: Areas smaller than 500~m².
\end{itemize}

An automated approach with manual refinement was used to create the reference data for training, validation, and testing. Initially, a mask was generated to filter nDSM values below a threshold of 3~m. This threshold was chosen to balance the exclusion of crops while including small trees and shrubs. Within this mask, the Normalized Difference Vegetation Index (NDVI) was calculated.
Because the NDVI histogram of elevated objects divides into active vegetation and the other impervious surfaces we applied an unsupervised k‑means clustering with $k=2$ to split the pixels into these two natural groups. This data-driven rule replaces a manually defined NDVI threshold. To fill small gaps within the trees mask, a morphological closing was performed using a rectangular structuring element with dimensions of $5\times5$ pixels, corresponding to a spatial size of $1\times1$~m based on the dataset’s spatial resolution. The results were then polygonized.
Since the earlier steps were pixel-based, they resulted in staircase-effects along polygon edges. To mitigate this and enhance the overall quality, the Douglas–Peucker algorithm was applied, simplifying and smoothing the edges while preserving the overall shape and maintaining important details \citep{Douglas_1973}.
To distinguish between forest and TOF, we follow the FAO definition, which classifies forests as areas larger than 0.5~ha and with a minimum width of 20~m~\citep{FAO_2013}. Polygon width was calculated using the shortest side of the smallest rotated enclosing rectangle. All polygons with a width bigger than 20~m and an area bigger than 0.5~ha were classified as \textit{Forests} and the remaining into the three TOF classes. Finally, a manual refinement step was conducted to ensure the quality of the reference data. All polygons were inspected at the native 0.2m scale against true‑colour orthophotos and the nDSM, and obvious errors were corrected. In particular, irregularly shaped polygons required adjustments; for example, \textit{Linear} objects that intersected a \textit{Forest} polygon were cut at the boundary and re‑classified. Every polygon was visually inspected at the native 0.2 m scale and corrected, producing the final reference layer.

Table~\ref{tab:object_geometry} summarizes the geometric properties of the reference objects per class. The classes are clearly separated along their defining dimensions, with \textit{Forest} the largest and \textit{Tree} the smallest, while \textit{Patch} and \textit{Linear} share a similar size and width range and differ mainly in elongation. This confirms that the chosen thresholds yield internally consistent classes.

\begin{table}[h]
\centering
\caption{Geometric properties of TOF classes in the reference data.}
\label{tab:object_geometry}
\resizebox{\textwidth}{!}{%
\begin{tabular}{lrrrrrr}
\toprule
 & \multicolumn{2}{c}{Area (m²)} & \multicolumn{2}{c}{Width (m)} & \multicolumn{2}{c}{Elongation} \\
\cmidrule(lr){2-3}\cmidrule(lr){4-5}\cmidrule(lr){6-7}
Class & Mean & Median & Mean & Median & Mean & Median \\
\midrule
Forest   & 103502.40 & 11435.58 & 188.54 & 111.98 & 2.03 & 1.76 \\
Patch    &    981.93 &   556.70 &  30.87 &  24.79 & 1.78 & 1.70 \\
Linear   &   1688.04 &   759.24 &  31.98 &  19.03 & 4.52 & 3.72 \\
Tree     &     44.44 &    20.24 &   5.90 &   4.71 & 1.51 & 1.28 \\
\bottomrule
\end{tabular}%
}
\end{table}

\begin{figure}[h]
    \centering
    \includegraphics[width=1\textwidth]{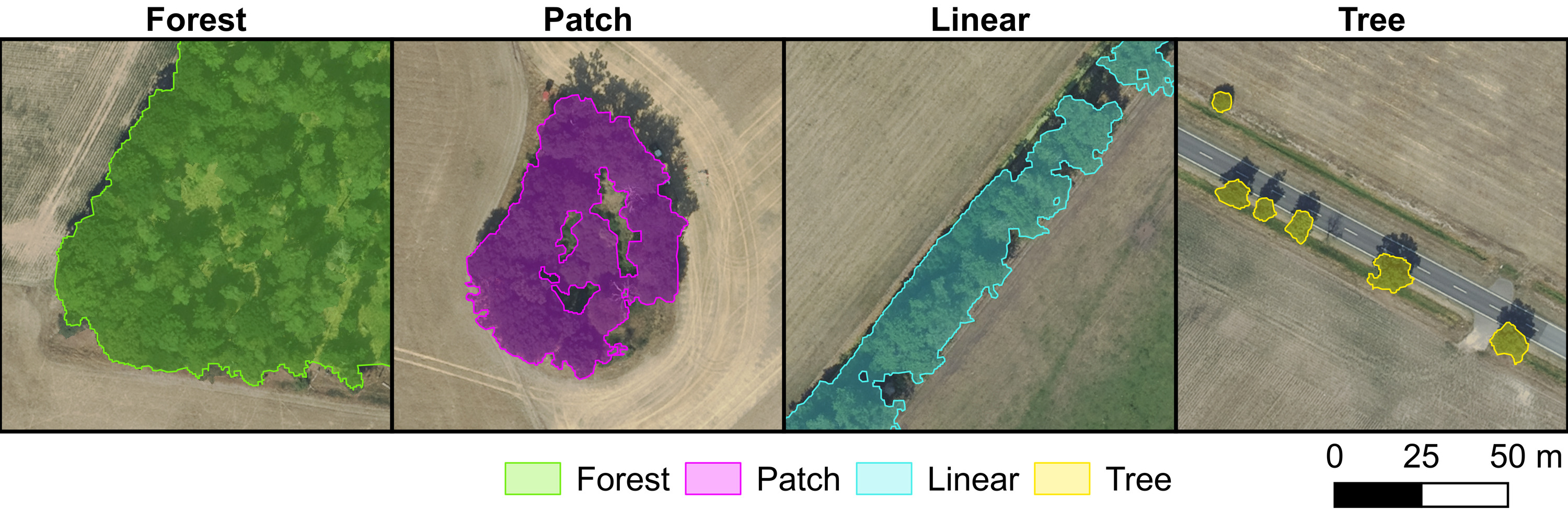}
    \caption{Examples of the structural classes: Forest, Patch, Linear, and Tree.}
    \label{fig:class_comparison}
\end{figure}

\subsection{Training, Validation, and Test Data}

All four study areas consist of 100 squared tiles of RGB aerial imagery, each covering an area of 1~km². The models were trained solely on RGB data, while the near-infrared band and nDSM were used exclusively for generating the reference data (Section~\ref{sec:generation_reference_data}).
A modality ablation (\ref{sec:modality_ablation}) shows that the NIR band contributes negligibly, whereas adding the nDSM improves accuracy by approximately 4\% mIoU. We nevertheless restrict the operational model to RGB, because RGB orthophotos are the most consistently available input across regions and acquisition years, whereas a co-registered image derived nDSM of comparable quality and date is usually not, which would hinder the large scale transfer.
Of the 100 tiles per study area, 90 were used for training, while five were allocated for validation and five for testing.
To ensure that all subsets were representative of the entire dataset, we calculated the class distribution for each study area (Table~\ref{tab:dataset_metadata}). The validation and test sets were selected by randomly choosing five tiles each, ensuring that their class proportions closely matched those of the whole study area, with a maximum deviation of one percent.

As these tiles ($5000\times5000$~pixels) are memory-consuming during the training process, they were split into patches of $1024\times1024$~pixels without overlap. During this process, the training tiles were augmented by horizontal and vertical flipping and added to the original patches. This resulted in a total of 27,000 training patches, including 9,000 non augmented ones, along with 500 validation patches, all evenly distributed across the study areas. The testing tiles were not split and were processed differently during inference. For details, please refer to Section~\ref{sec:postprocessing}.

\subsection{Model Selection}

As the first systematic evaluation of models for multiclass TOF mapping, the experiments and model selection were designed to provide an overview of the applicability of different architectural designs in TOF mapping and classification.
Accordingly, we selected six state-of-the-art models that have demonstrated strong accuracies in related remote sensing tasks on high-resolution benchmarks such as ISPRS Potsdam and Vaihingen \citepalias{ISPRSUrbanSemLab}: ABCNet, BANet, LSKNet, DC-Swin, and FT-UNetFormer, each representing a unique approach to feature extraction and context modeling. Additionally, we included the widely used U-Net as a baseline given its established role in environmental remote sensing applications \citep{Zhu_2017}. A summary of the models' architectures and parameters is provided in Table~\ref{tab:model_comparison}. This selection allows us to assess individual strengths across architectures. These range from bilateral and dual-path CNNs to advanced transformer-based architectures. Moreover, the chosen models are well-documented and supported by robust codebases, ensuring reproducibility. The fundamental encoder-decoder architecture of the models remained consistent with the configuration used in the benchmark dataset experiments.

\begin{table}[ht]
    \small
    \centering
    \caption{Comparison of the tested models, including the number of parameters and a brief description of their architectures.}
    \begin{tabular}{llp{6.6cm}}
        \toprule
        \textbf{Model} & \textbf{Parameters} & \textbf{Architecture}                                            \\
        \midrule
        ABCNet         & 13.6 M              & Bilateral CNN with spatial and contextual paths  \citep{li_2021}           \\
        BANet          & 12.7 M              & Parallel Transformer and CNN paths               \citep{wang_2021}        \\
        DC-Swin        & 66.9 M              & Swin Transformer encoder with dense decoder      \citep{wang_2022_2}      \\
        FT-UNetFormer  & 96.0 M              & Full Transformer with Swin encoder               \citep{wang_2022}        \\
        LSKNet         & 14.4 M              & CNN with Large Selective Kernel modules          \citep{li_2023}          \\
        U-Net          & 44.6 M              & Encoder-decoder CNN with skip connections         \citep{ronneberger_2015} \\
        \bottomrule
    \end{tabular}
    \label{tab:model_comparison}
\end{table}

\subsection{Training Setup}
\label{sec:training_setup}

\begin{table}[!]
\centering
\setlength{\tabcolsep}{4pt}
\caption{Training configuration and data-augmentation settings.}
\label{tab:train-config}
\begin{tabular}{@{}l l l@{}}
\toprule
\textbf{Section} & \textbf{Hyper-parameter} & \textbf{Value / Setting} \\ \midrule
\multirow{2}{*}{\textbf{Hardware}} %
    & GPUs                 & 4 × NVIDIA A100 (40 GB) \\
    & Iterations / epoch   & 6 750 \\ \midrule
\multirow{3}{*}{\textbf{Augment}} %
    & Photometric             & \makecell[l]{Random brightness, contrast\\saturation, hue} \\[2pt]
    & Geometric               & H/V flip, rotation (±15°) \\
    & Regularisation          & Coarse dropout \\ \midrule
\multirow{4}{*}{\textbf{Optimise}} %
    & Optimiser            & AdamW + Lookahead \\
    & Base LR / decay      & $6\times10^{-4}$ / $10^{-3}$ \\
    & Backbone LR / decay  & $1\times10^{-5}$ / $10^{-3}$ \\
    & Scheduler            & Cosine-annealing warm restarts \\ \midrule
\multirow{4}{*}{\textbf{Loss \& Stop}} %
    & Loss                 & Soft CE + Dice ($\alpha=0.05$) \\
    & Batch size           & 8 (4 for FT-UNetFormer) \\
    & Early stopping       & No mIoU gain for 3 epochs \\
    & Model selection      & Best validation mIoU \\ \bottomrule
\end{tabular}
\end{table}

All models were trained on four NVIDIA A100-SXM4 40 GB GPUs. The full set of data-augmentation steps, optimization hyper-parameters, and stopping criteria is summarized in Table \ref{tab:train-config}. After augmentation, a per-channel normalization was applied; the same normalization was used for the validation and test data to maintain consistency.
A batch size of 8 was used for every architecture except FT-UNetFormer, which was limited to 4 by GPU-memory constraints. While hyper-parameter optimization could further improve training efficiency and model accuracy, initial experiments showed only marginal gains. Training was stopped early if the validation mIoU failed to improve for three consecutive epochs, which occurred within 9 to 17 epochs depending on the architecture (More detail in Figure \ref{fig:training}). This criterion was chosen in light of the large number of iterations per epoch, and the checkpoint with the highest validation mIoU was selected for inference.

\subsection{Inference and Postprocessing}
\label{sec:postprocessing}

Instead of pre-cutting the original $5000\times5000$ tiles into smaller patches, inference was performed directly on the full tiles. To maintain the required $1024\times1024$ patch size for model input, a sliding window approach was applied. Specifically, patches of size $1024\times1024$ were extracted from the original $5000\times5000$ images, and their softmax probability predictions were stored. The sliding window was moved with a stride of 128, providing an optimal balance between inference time and prediction quality while minimizing the computation time associated with additional predictions.
To obtain the final result, overlapping predictions were merged by averaging the stored softmax probabilities, followed by majority voting for class assignment.
By processing entire tiles and considering overlapping patches, this approach not only improved segmentation and classification accuracy at patch boundaries by adding more spatial context to all predictions but also ensured the robustness of this approach for large-scale applications.

\subsection{Accuracy Assessment}

The accuracy was evaluated within 20 representative testing tiles, each with $5000\times5000$~pixels, evenly distributed across all study areas (Figure~\ref{fig:study_areas}). These tiles were not used during training. All pixels within the tiles were used to compute precision, recall, F1 score, and IoU. To compare the models, we calculated mIoU and mean F1 (mF1) score across all study areas using an unweighted mean across classes. This approach places greater emphasis on the smaller TOF classes, ensuring that evaluation metrics are more reflective of their accuracy. To further analyze the best-performing model, we computed an error matrix, area bias and the aforementioned metrics were calculated separately for each study area and class. We also compared it to the SWF Layer of 2015 to set our results into the context of existing products. We need to use the 2015 version as newer versions of the products dropped the structural classification scheme.

\subsection{Spatial Generalization}

To evaluate the spatial generalization of the method in unseen regions, the best-performing model was selected and trained on three study areas (Table~\ref{tab:transfer_datasets}). These areas were used for training and validation, while the remaining area was reserved for testing. This setup ensured that the model had no prior exposure to the landscape structure of the test site. The model was trained using the same hyperparameters as in previous experiments. The results were compared with those of the model trained on all study areas to assess spatial generalization capabilities.

\begin{table}[h]
    \centering
    \caption{Dataset configurations used to evaluate the spatial generalization capability of the method, with different study areas assigned to training, validation, and testing.}
    \begin{tabular}{lll}
        \toprule
        Dataset       & Training, Validation & Test   \\
        \midrule
        Combination 1 & SH, NRW\_N, NRW\_S   & BB
        \\
        Combination 2 & SH, BB, NRW\_S       & NRW\_N \\
        Combination 3 & SH, NRW\_N, BB       & NRW\_S \\
        Combination 4 & BB, NRW\_N, NRW\_S   & SH     \\
        \bottomrule
    \end{tabular}

    \label{tab:transfer_datasets}
\end{table}

\clearpage

\begin{landscape}

\begin{figure}[H]
    \centering
    \includegraphics[width=0.86\textwidth]{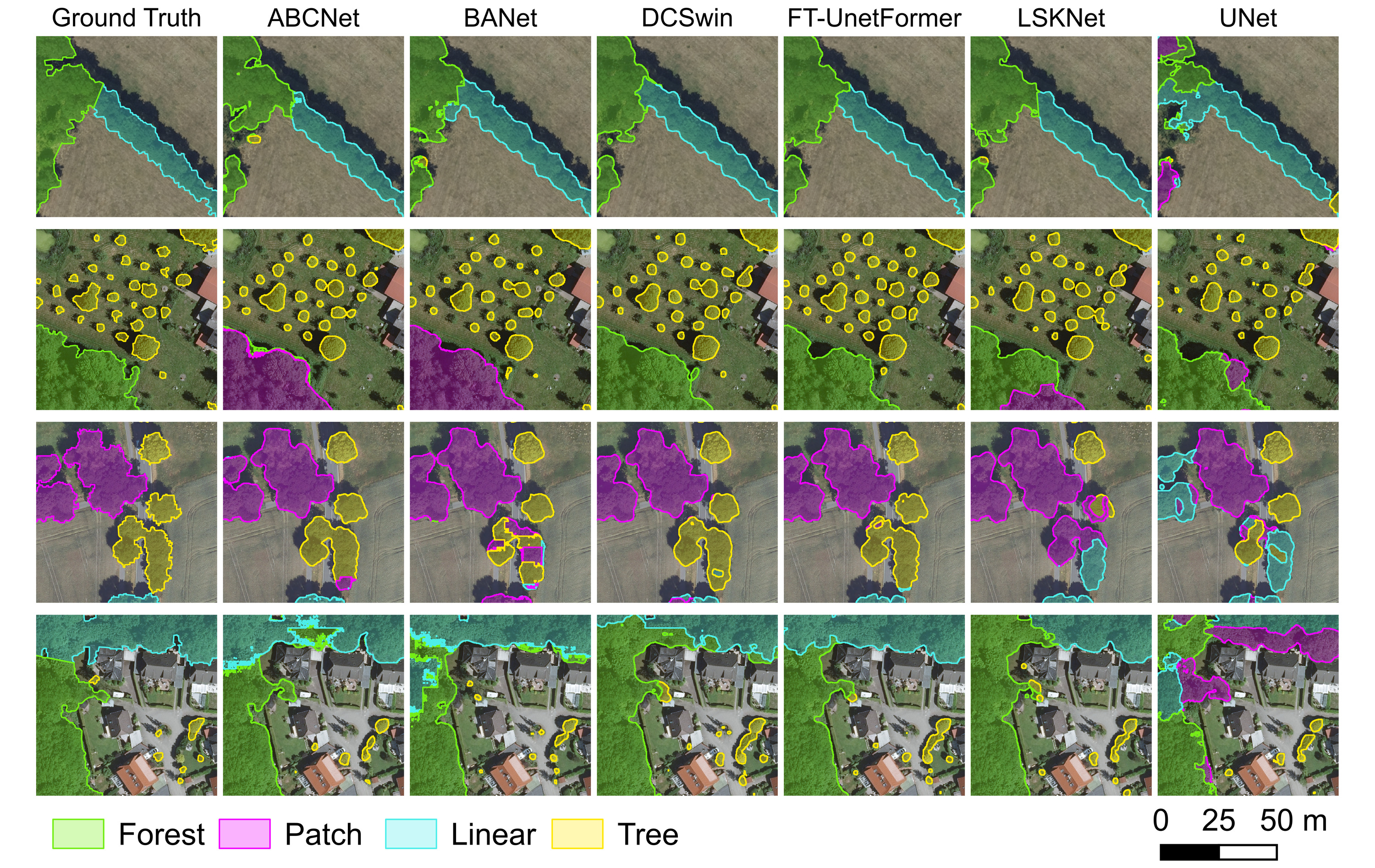}
    \caption{Comparison of TOF mapping and classification results for all tested models in selected subregions.}
    \label{fig:model_comparison}
\end{figure}

\begin{table}[!ht]
    \centering
    \caption{Per-class IoU and F1 scores across all study areas. mIoU and mF1 are calculated as unweighted averages over all classes. Standard deviations indicated by ($\pm$) show variation across the four study areas. The highest values for each metric are highlighted in bold.}
    \resizebox{\linewidth}{!}{%
        \begin{tabular}{llllllllllllll}\toprule
                                                & \multicolumn{2}{c}{Mean}                             & \multicolumn{2}{c}{Forest}                          & \multicolumn{2}{c}{Patch}                            & \multicolumn{2}{c}{Linear}                          & \multicolumn{2}{c}{Tree}
                                                \\ \cmidrule(lr){2-3} \cmidrule(lr){4-5} \cmidrule(lr){6-7} \cmidrule(lr){8-9} \cmidrule(lr){10-11} \cmidrule(lr){12-13}
        Model                                    & mIoU                      & mF1                      & IoU                      & F1                       & IoU                      & F1                        & IoU                      & F1                       & IoU                      & F1                        \\ \midrule
        ABCNet                                   & 0.709 $\pm$ 0.037           & 0.821 $\pm$ 0.025         &  0.944 $\pm$ 0.032          & 0.971 $\pm$ 0.058 & 0.532 $\pm$ 0.067          & 0.695 $\pm$ 0.056           & 0.739 $\pm$ 0.063          & 0.850 $\pm$ 0.042          & 0.622 $\pm$ 0.011          & 0.767 $\pm$ 0.009           \\
        BANet                                    & 0.656 $\pm$ 0.025           & 0.777 $\pm$ 0.022         & 0.937 $\pm$ 0.058          & 0.967 $\pm$ 0.032          & 0.422 $\pm$ 0.081          & 0.593 $\pm$ 0.082           & 0.689 $\pm$ 0.027          & 0.816 $\pm$ 0.019          & 0.575 $\pm$ 0.025          & 0.730 $\pm$ 0.021           \\
        DC-Swin                                   & 0.714 $\pm$ 0.019           & 0.824 $\pm$ 0.018         & 0.949 $\pm$ 0.037          & 0.974 $\pm$ 0.020          & 0.570  $\pm$ 0.075         & 0.726 $\pm$ 0.065           & 0.751  $\pm$ 0.017         & 0.858 $\pm$ 0.011          & 0.586 $\pm$ 0.023          & 0.727 $\pm$ 0.019           \\
        FT-UNetFormer                            & \textbf{0.739} $\pm$ 0.016  &\textbf{0.843} $\pm$ 0.012 & \textbf{0.952} $\pm$ 0.045          & \textbf{0.975} $\pm$ 0.024 & \textbf{0.606} $\pm$ 0.055 & \textbf{0.754} $\pm$ 0.044  & \textbf{0.774} $\pm$ 0.020 & \textbf{0.872} $\pm$ 0.013 & \textbf{0.626} $\pm$ 0.019 & \textbf{0.770} $\pm$ 0.015  \\
        LSKNet                                   & 0.697 $\pm$ 0.016           & 0.812 $\pm$ 0.013         & 0.943 $\pm$ 0.047          & 0.971 $\pm$ 0.026          & 0.545 $\pm$ 0.058          & 0.705 $\pm$ 0.052           & 0.718 $\pm$ 0.030          & 0.836 $\pm$ 0.020          & 0.582 $\pm$ 0.016          & 0.736 $\pm$ 0.013           \\
        U-Net                                    & 0.484 $\pm$ 0.049           & 0.618 $\pm$ 0.050         & 0.804 $\pm$ 0.103          & 0.891 $\pm$  0.065         & 0.151 $\pm$ 0.061          & 0.263 $\pm$ 0.095           & 0.527 $\pm$ 0.049          & 0.690 $\pm$ 0.043          & 0.456 $\pm$ 0.065          & 0.626 $\pm$ 0.063           \\
    \bottomrule
    \end{tabular}
    }

    \label{tab:model_comparison_results}
\end{table}

\end{landscape}

\section{Results}

\subsection{Model Comparison}

Figure~\ref{fig:model_comparison} and Table~\ref{tab:model_comparison_results} present the mapping and classification results for all models. FT-UNetFormer achieved the best mean accuracy (mIoU 0.739, mF1 0.843). The other models performed competitively, while the baseline U-Net was clearly the weakest. FT-UNetFormer was also among the most consistent models across the four study areas, whereas ABCNet, BANet, and U-Net varied more, as reflected in Figure~\ref{fig:model_comparison}.
Across classes, all models classified the \textit{Forest} class with high accuracy and differed mainly on the more complex TOF classes. Among these, \textit{Linear} reached the highest accuracy and \textit{Patch} the lowest, with \textit{Tree} in between. \textit{Patch} also showed the largest variability between models. FT-UNetFormer achieved the best result in every class and was therefore selected for the detailed analysis.

\subsection{Detailed Results and Error Analysis}

\begin{table}[h]
    \centering
    \caption{Normalized error matrix for the best-performing model (FT-UNetFormer). Rows represent ground truth classes. Columns show the predicted classification results.}
    \begin{tabular}{llllll}
        \toprule
        \multicolumn{1}{c}{\textbf{Ground Truth}} & \multicolumn{5}{c}{\textbf{Classification Results}}                                   \\
        \cmidrule(lr){1-1} \cmidrule(lr){2-6}
                                                  & Background                                          & Forest & Patch & Linear & Tree  \\
        \midrule
        Background                                & 99.17                                               & 0.37   & 0.07  & 0.24   & 0.16  \\
        Forest                                    & 1.34                                                & 97.94  & 0.23  & 0.40   & 0.09  \\
        Patch                                     & 5.61                                                & 8.11   & 73.19 & 9.16   & 3.93  \\
        Linear                                    & 5.05                                                & 4.70   & 1.37  & 86.34  & 2.54  \\
        Tree                                      & 11.78                                               & 1.65   & 3.27  & 5.01   & 78.28 \\
        \bottomrule
    \end{tabular}
    \label{tab:error_matrix}
\end{table}

\begin{figure}[!]
    \centering
    \includegraphics[width=1\textwidth]{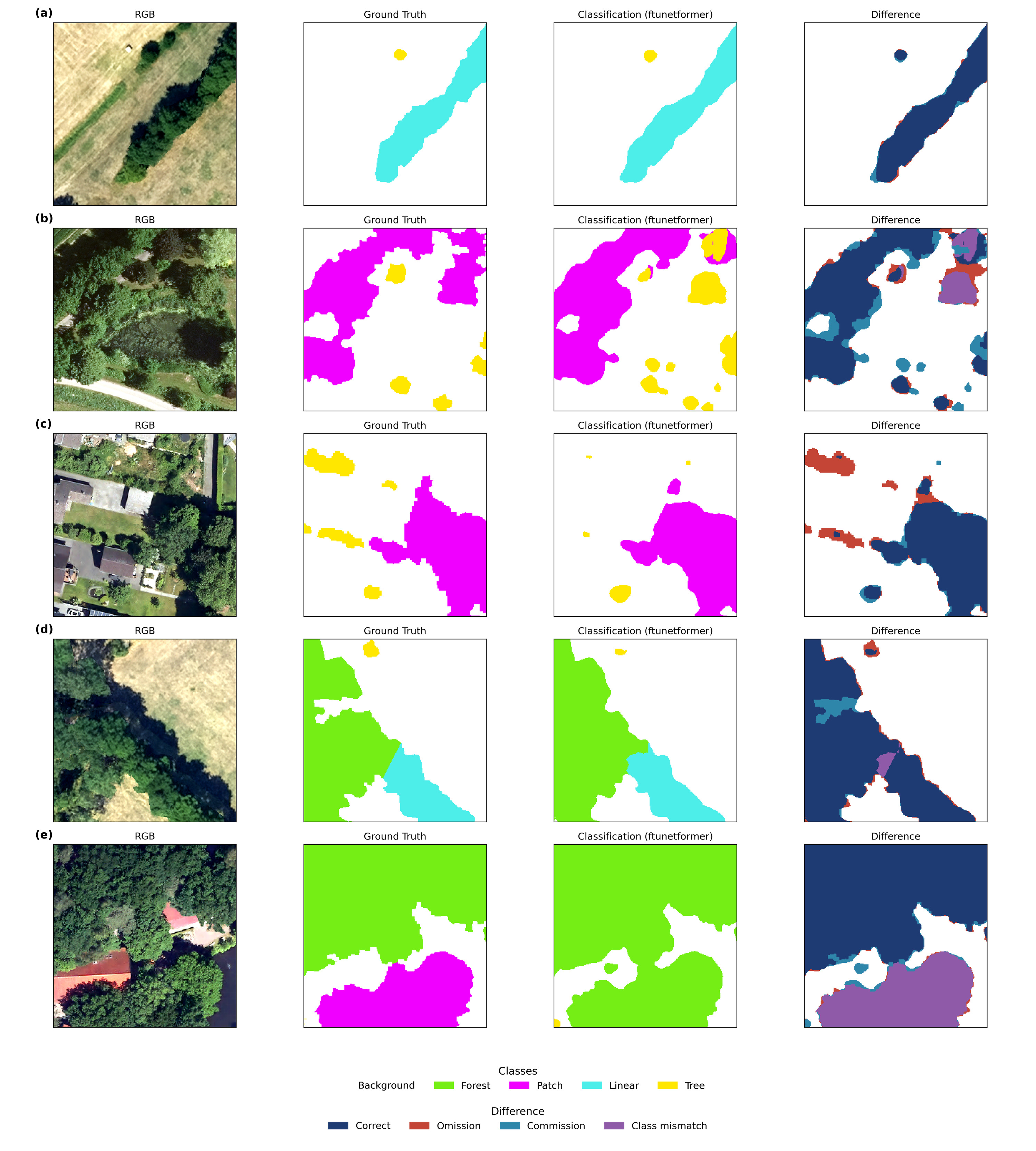}
    \caption{Qualitative results of FT-UNetFormer on selected examples (a–e), showing the RGB orthophoto, ground truth, classification, and a difference map. The difference map marks each pixel as Correct, Omission, Commission, or Class mismatch.}
    \label{fig:detailed_examples}
\end{figure}

FT-UNetFormer delineated the \textit{Forest} and TOF classes accurately (Figure~\ref{fig:detailed_examples} and Figure~\ref{fig:best_model_detail} for contextual results). \textit{Forest} was classified most accurately, followed by the spatially more complex TOF classes, among which \textit{Linear} was the most accurate and \textit{Patch} the least (Table~\ref{tab:model_comparison_results}).
The error matrix (Table~\ref{tab:error_matrix}) shows three main sources of misclassification. First, edge pixels are the primary challenge for the TOF classes, most pronounced for \textit{Tree}, which contains the most edge pixels and is most often confused with \textit{Background} (Figure~\ref{fig:detailed_examples}a,b,c). Second, \textit{Patch} and \textit{Linear} are confused with \textit{Forest} where larger patches and hedgerows merge into adjacent forest (Figure~\ref{fig:detailed_examples}d, e). Third, the TOF classes are confused with each other, most often \textit{Patch} predicted as \textit{Linear} in transition zones, as well as confusion between \textit{Patch} and \textit{Tree} (Figure~\ref{fig:detailed_examples}b,c).
The area-based comparison (Table~\ref{tab:area_comparison}) confirms these patterns: \textit{Background} and \textit{Forest} areas are reproduced with low bias, while \textit{Linear} and especially \textit{Patch} are underestimated and \textit{Tree} is overestimated.

\label{chap:best_model_classes}

\begin{table}[h]
\centering
\caption{Area comparison between ground truth and prediction per class.}
\label{tab:area_comparison}
\begin{tabular}{lccc}
\toprule
Class & GT Area (ha) & Pred.\ Area (ha) & Bias (\%) \\
\midrule
Background & 1549.72 & 1548.02 & $-0.11$ \\
Forest &  347.97 &  359.40 & $+3.28$ \\
Patch &   22.95 &   15.30 & $-33.35$ \\
Linear &   70.07 &   64.20 & $-8.38$ \\
Tree &   17.16 &   20.95 & $+22.14$ \\
\bottomrule
\end{tabular}
\end{table}

\begin{table}[!b]
    \centering
    \caption{Comparison of IoU, F1, precision, and recall of FT-UNetFormer for different classes across study areas.}
        \begin{tabular}{llllll}
            \toprule
            Metric    & Class  & SH   & BB   & NRW\_N & NRW\_S \\
            \midrule
            IoU       & Forest & 0.97 & 0.96 & 0.88   & 0.96   \\
                      & Patch  & 0.59 & 0.51 & 0.61   & 0.64   \\
                      & Linear & 0.79 & 0.80 & 0.77   & 0.75   \\
                      & Tree   & 0.62 & 0.59 & 0.63   & 0.64   \\
            \midrule
            F1        & Forest & 0.99 & 0.98 & 0.93   & 0.98   \\
                      & Patch  & 0.74 & 0.68 & 0.76   & 0.78   \\
                      & Linear & 0.88 & 0.89 & 0.87   & 0.86   \\
                      & Tree   & 0.77 & 0.74 & 0.77   & 0.78   \\
            \midrule
            Precision & Forest & 0.98 & 0.98 & 0.92   & 0.97   \\
                      & Patch  & 0.82 & 0.68 & 0.73   & 0.82   \\
                      & Linear & 0.87 & 0.91 & 0.90   & 0.86   \\
                      & Tree   & 0.75 & 0.72 & 0.75   & 0.78   \\
            \midrule
            Recall    & Forest & 0.99 & 0.98 & 0.94   & 0.98   \\
                      & Patch  & 0.67 & 0.67 & 0.78   & 0.74   \\
                      & Linear & 0.89 & 0.87 & 0.84   & 0.85   \\
                      & Tree   & 0.79 & 0.77 & 0.79   & 0.77   \\
            \bottomrule
        \end{tabular}
    \label{tab:accuracy_study_areas}
\end{table}

Accuracy was relatively consistent across the four study areas (Table~\ref{tab:accuracy_study_areas}). \textit{Linear} and \textit{Tree} were the most consistent and \textit{Patch} the most variable, with its lowest accuracy in BB. Precision and recall were well balanced except for \textit{Patch}, which varied between regions, and \textit{Tree}, which consistently showed higher recall than precision, indicating overestimation through oversegmentation at object edges (Section~\ref{chap:best_model_classes}).

\begin{figure}[!h]
    \centering
    \includegraphics[width=1\textwidth]{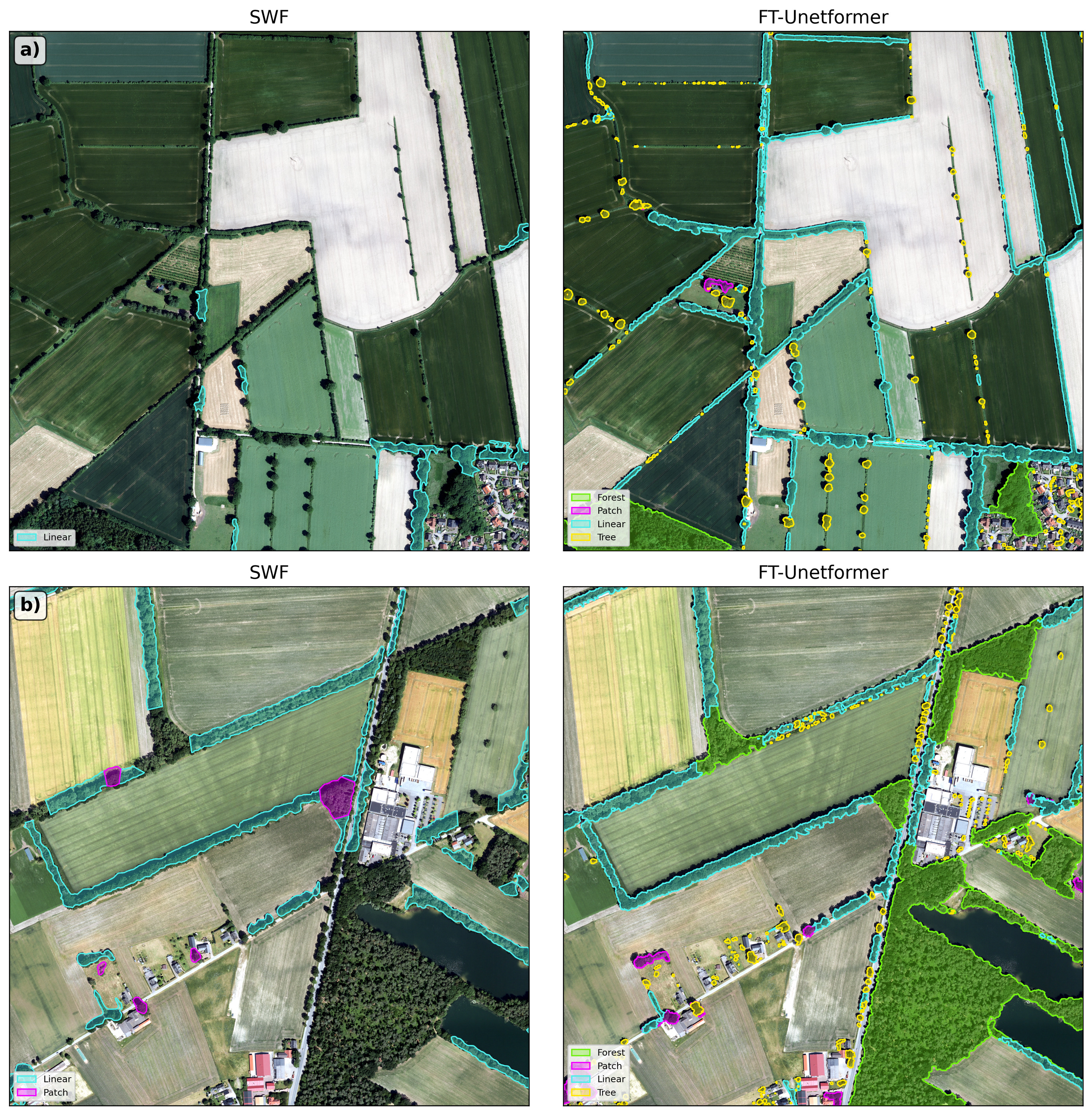}
    \caption{Comparison of the SWF (left) and the FT-UNetFormer prediction (right) over example areas a and b. SWF represents structural classes as \textit{Linear} and \textit{Patch} only, whereas FT-UNetFormer additionally maps individual \textit{Tree} objects and \textit{Forest}.}
    \label{fig:swf_comparison}
\end{figure}

\subsection{Comparison with SWF}

We compared the FT-UNetFormer prediction with the SWF for the \textit{Linear} and \textit{Patch} classes (Table~\ref{tab:swf_vs_pred}, Figure~\ref{fig:swf_comparison}). In Figure~\ref{fig:swf_comparison}a SWF captures only small parts of the woody features, while in row b SWF over-delineates the features at their boundaries. Overall, SWF maps a larger \textit{Linear} area, reflecting its coarser 5~m resolution. In comparison the FT-UNetFormer maps considerably more \textit{Patch} area and resolves numerous individual trees that SWF does not represent. The low spatial overlap (IoU of 26.9\% for \textit{Linear} and 5.0\% for \textit{Patch}) follows from these differences in resolution and class definition.

\begin{table}[!]
\centering
\caption{Area-based comparison of SWF and the prediction (ftunetformer) for the linear and patch classes, aggregated across all study areas.}
\label{tab:swf_vs_pred}
\begin{tabular}{lrrrrr}
\toprule
Class  & SWF (ha) & Pred.\ (ha) & Pred.\ vs SWF & Intersection (ha) & IoU \\
\midrule
Linear & 85.39    & 63.93       & $-25.1\%$     & 31.62             & $26.9\%$ \\
Patch  & 4.05     & 15.23       & $+276.0\%$    & 0.91              & $5.0\%$ \\
\bottomrule
\end{tabular}
\end{table}

\subsection{Spatial Generalization}

\begin{table}[!h]
    \centering
    \caption{Comparison of spatial generalization of the FT-UNetFormer. The table presents IoU, F1 scores, precision, and recall for different classes across study areas. Values in parentheses indicate the difference between the model trained on three study areas and the model trained on all study areas.}
    \resizebox{\textwidth}{!}{%
        \begin{tabular}{llllll}
            \toprule
            Metric    & Class  & SH           & BB           & NRW\_N       & NRW\_S       \\
            \midrule
            IoU       & Forest & 0.94 (-0.03) & 0.91 (-0.05) & 0.86 (-0.02) & 0.89 (-0.07) \\
                      & Patch  & 0.36 (-0.23) & 0.38 (-0.13) & 0.48 (-0.13) & 0.47 (-0.17) \\
                      & Linear & 0.71 (-0.08) & 0.59 (-0.21) & 0.74 (-0.03) & 0.53 (-0.22) \\
                      & Tree   & 0.44 (-0.18) & 0.46 (-0.13) & 0.52 (-0.11) & 0.47 (-0.17) \\
            \midrule
            F1        & Forest & 0.97 (-0.02) & 0.95 (-0.03) & 0.92 (-0.01) & 0.94 (-0.04) \\
                      & Patch  & 0.52 (-0.22) & 0.55 (-0.13) & 0.65 (-0.11) & 0.64 (-0.14) \\
                      & Linear & 0.83 (-0.05) & 0.74 (-0.15) & 0.85 (-0.02) & 0.69 (-0.17) \\
                      & Tree   & 0.61 (-0.16) & 0.63 (-0.11) & 0.68 (-0.09) & 0.64 (-0.14) \\
            \midrule
            Precision & Forest & 0.95 (-0.03) & 0.97 (-0.01) & 0.89 (-0.03) & 0.94 (-0.03) \\
                      & Patch  & 0.63 (-0.19) & 0.50 (-0.32) & 0.65 (-0.08) & 0.60 (-0.22) \\
                      & Linear & 0.78 (-0.09) & 0.96 (0.05)  & 0.85 (-0.05) & 0.95 (0.09)  \\
                      & Tree   & 0.79 (0.04)  & 0.62 (-0.10) & 0.69 (-0.06) & 0.56 (-0.19) \\
            \midrule
            Recall    & Forest & 0.99 (0.00)  & 0.94 (-0.04) & 0.96 (0.02)  & 0.92 (-0.06) \\
                      & Patch  & 0.45 (-0.22) & 0.61 (-0.06) & 0.65 (-0.13) & 0.68 (-0.06) \\
                      & Linear & 0.90 (0.01)  & 0.61 (-0.26) & 0.85 (0.01)  & 0.55 (-0.30) \\
                      & Tree   & 0.50 (-0.29) & 0.64 (-0.13) & 0.67 (-0.12) & 0.74 (-0.05) \\
            \bottomrule
        \end{tabular}
    }
    \label{tab:generalization_metrics}
\end{table}

\begin{figure}[!h]
    \centering
    \includegraphics[width=1\textwidth]{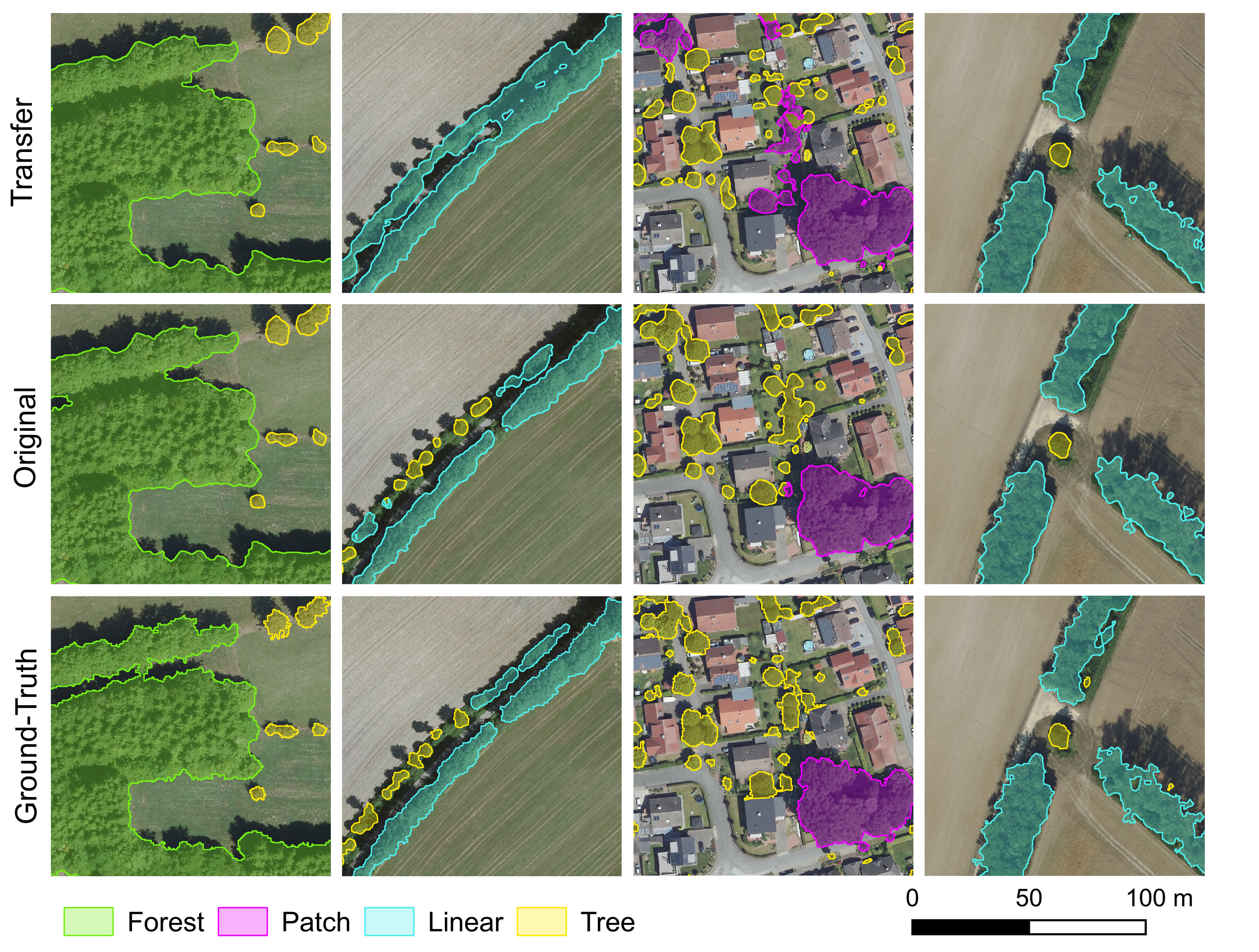}
    \caption{The first row shows results from the FT-UNetFormer trained on three study areas and tested on an unseen region. The second row presents results from the model trained on all study areas. The third row displays the ground truth.}

    \label{fig:generalization_results}
\end{figure}

Table~\ref{tab:generalization_metrics} presents the change in accuracy when FT-UNetFormer was trained on three study areas and tested on the remaining one (Table~\ref{tab:transfer_datasets}). Accuracy decreased across all classes and regions.
The \textit{Forest} class generalized best and remained stable across regions, consistent with its well-defined boundaries in Figure~\ref{fig:generalization_results}. The TOF classes degraded more strongly. \textit{Patch} was affected most across all metrics and regions, and \textit{Linear} declined sharply in NRW\_S and BB, where recall dropped while precision was maintained or improved. \textit{Tree} also degraded, without a clear regional trend.

\clearpage

\section{Discussion}

Our study demonstrates that deep learning, particularly transformer-based architectures, enables accurate mapping and classification of TOF across agricultural landscapes. By applying our approach to four distinct regions, we extend the analysis beyond localized studies and show its potential for large-scale application.

\subsection{Comparison of Model Architectures}

FT-UNetFormer achieved the best and most consistent accuracy across all classes (Table~\ref{tab:model_comparison_results}, Figure~\ref{fig:model_comparison}). The models differed little on the \textit{Forest} and \textit{Linear} classes and diverged mainly on the harder \textit{Patch} and \textit{Tree} classes, indicating that the choice of architecture matters most where spatial complexity is highest.
This is reflected in the architectural design. Models with advanced spatial-context mechanisms (ABCNet, BANet, LSKNet, DC-Swin, and FT-UNetFormer) handled the complexity of the TOF classes better than the conventional U-Net, whose symmetric encoder-decoder with skip connections \citep{ronneberger_2015} captures spatial context less effectively (Figure~\ref{fig:model_comparison}). Transformer-based architectures outperformed their CNN-based counterparts and were more consistent across study areas. Among the transformer-based models, the fully transformer-based FT-UNetFormer outperformed DC-Swin, which uses a CNN decoder \citep{wang_2022, wang_2022_2}: by replacing the CNN encoder with a Swin Transformer, both encoding and decoding benefit from self-attention, in line with the strong performance of transformers on context-dependent remote sensing tasks \citep{Aleissaee_2023}. BANet, however, performed poorest among the spatial-context models despite its transformer dependency path, suggesting that integrating transformer components is not sufficient on its own and that overall model complexity also matters. Consistently, the two models with the most parameters, FT-UNetFormer and DC-Swin, achieved the highest accuracy.
Earlier deep learning studies of TOF mainly addressed binary classification (TOF versus non-TOF) and reached high accuracy with a U-Net, for example \citet{Vinod_2023} (F1 0.93), \citet{Reiner_2023} (R² 0.62 to 0.93), and \citet{Liu_2023} (IoU 0.69). In the multiclass setting, however, the U-Net performed clearly worse than the spatial-context CNNs and transformers. This indicates that transformer-based architectures are particularly suited to the higher spatial complexity of multiclass TOF mapping and are a promising direction for further development.

\subsection{Detailed Results of the Classification}
FT-UNetFormer reached high overall accuracy (IoU 0.74, F1 0.84), although accuracy varied by class (Tables~\ref{tab:error_matrix},~\ref{tab:accuracy_study_areas}). Earlier TOF studies differ in sensor, region, and class focus, so the following comparison is indicative rather than direct, but it reveals consistent trends.
Despite addressing the harder multiclass problem, our accuracy is competitive with simpler products. For the structural classes (\textit{Linear} and \textit{Patch}), it exceeds the  SWF evaluated on the same structures \citep{CopernicusSWF_2024} and recent large-area binary TOF maps \citep{Brandt_2024, Liu_2023}. This shows that a transferable multiclass classification can be achieved without sacrificing accuracy relative to coarser, single-class products.
The accuracy ranking is consistent across classes and studies: \textit{Forest} is classified best, followed by \textit{Linear}, \textit{Tree}, and \textit{Patch}. The same pattern appears in all TOF studies regardless of method \citep{Straub_2008, Bolyn_2019, MaurizioSarti_2021}, indicating that the difficulty of TOF mapping is intrinsic to the structures, with their high edge density and spatial complexity, rather than to a particular model.
The \textit{Forest} class is the easiest, because large continuous canopies have little edge relative to their area. Our results approach the near-perfect rule-based values of \citet{Straub_2008} while covering a far larger area, and clearly exceed medium-resolution mapping \citep{MaurizioSarti_2021}. The few errors occur at fuzzy forest edges and undetected clearings, as also reported earlier \citep{MaurizioSarti_2021, Straub_2008}.
The \textit{Patch} class is the most difficult, reflecting its position as a transitional category between \textit{Forest}, \textit{Linear}, and \textit{Tree}. Our accuracy is comparable to rule-based approaches on small test areas \citep{Straub_2008, Bolyn_2019} and far above medium-resolution mapping \citep{MaurizioSarti_2021}. The higher precision of \citet{Bolyn_2019} stems from their small study area and manually tuned thresholds rather than a transferable method, which highlights the central trade-off of our approach. It accepts lower accuracy on the hardest class in exchange for transferable, tuning-free mapping across landscapes. To test whether class balancing could reduce this limitation, we applied class-aware oversampling of \textit{Patch} tiles (\ref{sec:oversampling}). The class-aware sampling improved the detection of the \textit{Patch} class, but the gain in recall came at the cost of precision, so the net gains in IoU and F1 remained modest. The \textit{Forest} class lost recall at constant precision, which indicates that part of the additional \textit{Patch} predictions originated from forest boundaries, where the two classes are structurally ambiguous. The sampling therefore shifted the model toward a less conservative treatment of small tree structures rather than improving class separability. It is suitable when rare structural classes are the priority and a moderate loss in \textit{Forest} recall is acceptable, whereas applications that require accurate forest extent should apply it with caution.
For the \textit{Linear} class, our accuracy equals the rule-based pipeline of \citet{Bolyn_2019} but is achieved across several study sites rather than one, showing that deep learning can match manually tuned pipelines while scaling. Dedicated hedgerow studies report lower accuracy \citep{Huber-Garcia_2025, Muro_2025}, which suggests that large-scale linear-feature mapping remains challenging and that transformer-based models are advantageous. Errors arise mainly where hedgerows merge into adjacent forest, producing ambiguous boundaries (Figure~\ref{fig:detailed_examples}).
The \textit{Tree} class remains difficult. Our recall improves on older work \citep{Straub_2008} and matches \citet{Pujar_2014}, while precision is slightly below the rule-based value of \citet{Bolyn_2019}. Most errors occur at crown edges that blur into background or hedgerows, as also noted by \citet{Bolyn_2019}. Soft-boundary approaches, such as the heatmap-based classification of \citet{Brandt_2024}, treat crown edges as gradual rather than strict and are a promising direction for reducing these edge errors.
Finally, accuracy was stable across study areas for most classes, but the \textit{Patch} class varied more (Table~\ref{tab:accuracy_study_areas}), most likely because too few examples represent its structural variability, whereas the well-represented \textit{Linear} and \textit{Tree} classes generalized more consistently. This points to class imbalance as a key limitation, which could be addressed by oversampling the TOF classes while undersampling \textit{Forest} and \textit{Background}, or by augmenting the training data with synthetic examples.
More generally, edge and boundary errors are a known limitation of segmentation models trained with pixel-wise losses, and several approaches address them. Incorporating height models where available is one option, as the modality ablation indicates (\ref{sec:modality_ablation}). Boundary-aware losses such as the boundary loss \citep{kervadec_2019} or Lovász-Softmax \citep{berman_2018} upweight boundary and small-object pixels. Edge-preserving architectures such as Gated-SCNN \citep{takikawa_2019} and PointRend \citep{kirillov_2020} preserve sharper object outlines. Post-processing methods such as conditional random fields \citep{krahenbuhl_2011} or minimum mapping unit filtering reduce fragmentation. These approaches offer concrete directions for improving the delineation of small and fragmented TOF objects.

\subsection{Generalization Capabilities}
As expected, accuracy declined for all classes, when spatially transferring the model (Table~\ref{tab:generalization_metrics}). The decline was smallest for \textit{Forest} and largest for the TOF classes, with \textit{Patch} most affected and overestimated in BB and NRW\_S.
The decline can be attributed to landscape structure as the main factor. The model transferred best between SH and NRW\_N, which, despite being hundreds of kilometers apart, share the geomorphology and traditional smallholder agriculture of the north German plains \citep{Meynen_1962}, and therefore similar field boundaries and linear vegetation. Transfer was weaker to BB, characterized by collectivized agriculture with large plots, and to NRW\_S, a hilly pasture region with extensive forests \citep{Wolff_2021}. \citet{Muro_2025} reported the same effect for hedgerow detection, with accuracy decreasing when a model trained in northern Germany was applied in the south, and likewise attributed it to differing landscape structure.
These results show that the model generalizes well for the \textit{Forest} class but requires training data from structurally similar landscapes to classify the TOF classes reliably \citep{chadwick_2024}. To our knowledge, no other TOF classification study has systematically tested transfer across diverse agricultural landscapes. Binary TOF studies report comparable regional variation, however; \citet{Liu_2023}, for example, found lower accuracy in Mediterranean than in boreal and temperate regions, attributed to training data and landscape differences, consistent with our findings.
While we address domain shift here by recommending more diverse training data, several methodological developments target this problem more directly. Domain adaptation and domain generalization aim to transfer models to unseen regions with little or no labeling of the target domain \citep{tuia_2016}. Self-supervised pretraining on large amounts of unlabeled imagery learns transferable representations and reduces the dependence on labeled data \citep{wang_ssl_2022}, and large-scale geospatial foundation models extend this idea to continental and global pretraining \citep{reed_2023}. Applying such approaches is a promising route to improving cross-landscape transfer in future TOF mapping.

\subsection{Validation and Uncertainty}

Several sources of uncertainty affect the interpretation of these results. The reference data were generated semi-automatically from height, NDVI, and clustering, followed by manual refinement. The manual step removes obvious errors, but some uncertainty remains, particularly at object boundaries and for objects close to the defining class thresholds. Because the classes are delimited by fixed area, width, and elongation thresholds, objects in the transition zones between classes, such as a woody patch near the 0.5 ha Forest boundary or a compact grove near the Patch-Linear elongation limit, are inherently ambiguous. Part of the confusion between neighbouring classes, especially for the \textit{Patch} class, therefore reflects this reference and definition uncertainty rather than model error alone. Consistently, the misclassifications concentrate at object edges and in transition zones rather than occurring randomly (Table~\ref{tab:error_matrix}, Figure~\ref{fig:detailed_examples}). For practical applications this means that the total woody cover and the broad distinction between forest and TOF are robust, whereas the precise partition between adjacent structural classes is subject to a higher uncertainty and should be interpreted accordingly.

\section{Conclusion and Outlook}

In this study, we evaluated six CNN-, transformer-, and hybrid-based semantic segmentation models for mapping and classifying TOF in four agricultural landscapes in Germany. The FT-UNetFormer achieved the best performance (mIoU 0.74; mF1 0.84), underscoring the importance of context understanding through global–local feature modeling. The model achieved accurate results for all TOF classes, with IoU(F1) scores of 0.97(0.98) for the \textit{Forest} class, 0.61(0.75) for \textit{Patch}, 0.77(0.87) for \textit{Linear}, and 0.63(0.77) for \textit{Tree}. Despite these promising results, the spatial complexity and higher edge density of TOF structures, particularly for the \textit{Patch} class, present challenges. To evaluate the model's ability for large-scale mapping within Germany, its spatial generalization was tested on previously unseen agricultural landscapes. Although the model shows adequate generalization capability, our findings suggest that for large-scale mapping, training data from regions with similar landscape characteristics is essential.

Future studies should investigate methods that exploit larger spatial context, such as vision transformers, for TOF mapping and classification. Adding the height information to the input data improves accuracy, a flexible architecture that fuses image with height information when available, would combine both advantages. Training strategies such as oversampling also merit investigation to address the classification challenges of the TOF classes. The approach can further be adapted to high-resolution satellite imagery, enabling scalable TOF mapping in regions without aerial imagery. Ultimately, the framework provides information that supports policymakers in land-use planning and in promoting agroforestry.

\section*{CRediT authorship contribution statement}
\textbf{Moritz Lucas:} Writing -- original draft, Visualization, Validation, Methodology, Investigation, Formal analysis, Data curation, Conceptualization. \textbf{Hamid Ebrahimy:} Writing -- review \& editing, Supervision, Project administration, Conceptualization. \textbf{Viacheslav Barkov:} Writing -- review \& editing, Software. \textbf{Ralf Pecenka:} Funding acquisition, Project administration, Supervision. \textbf{Kai-Uwe Kühnberger:} Writing -- review \& editing, Project administration, Funding acquisition. \textbf{Björn Waske:} Writing -- review \& editing, Supervision, Project administration, Funding acquisition, Conceptualization.

\section*{Declaration of competing interest}
The authors declare that they have no known competing financial interests or personal relationships that could have appeared to influence the work reported in this paper.

\section*{Acknowledgments}
This research was supported by the Lower Saxony Ministry of Science and Culture (MWK), funded through the zukunft.niedersachsen program of the Volkswagen Foundation.

\section*{Data availability}
The dataset is openly available on Zenodo \citep{Lucas_2025_zenodo}, and the code is available at \href{https://github.com/Moerizzy/TOFMapper}{https://github.com/Moerizzy/TOFMapper}.

\clearpage

\appendix

\section{}

\subsection{Model Training}

\begin{figure}[ht]
    \centering
    \includegraphics[width=1\textwidth]{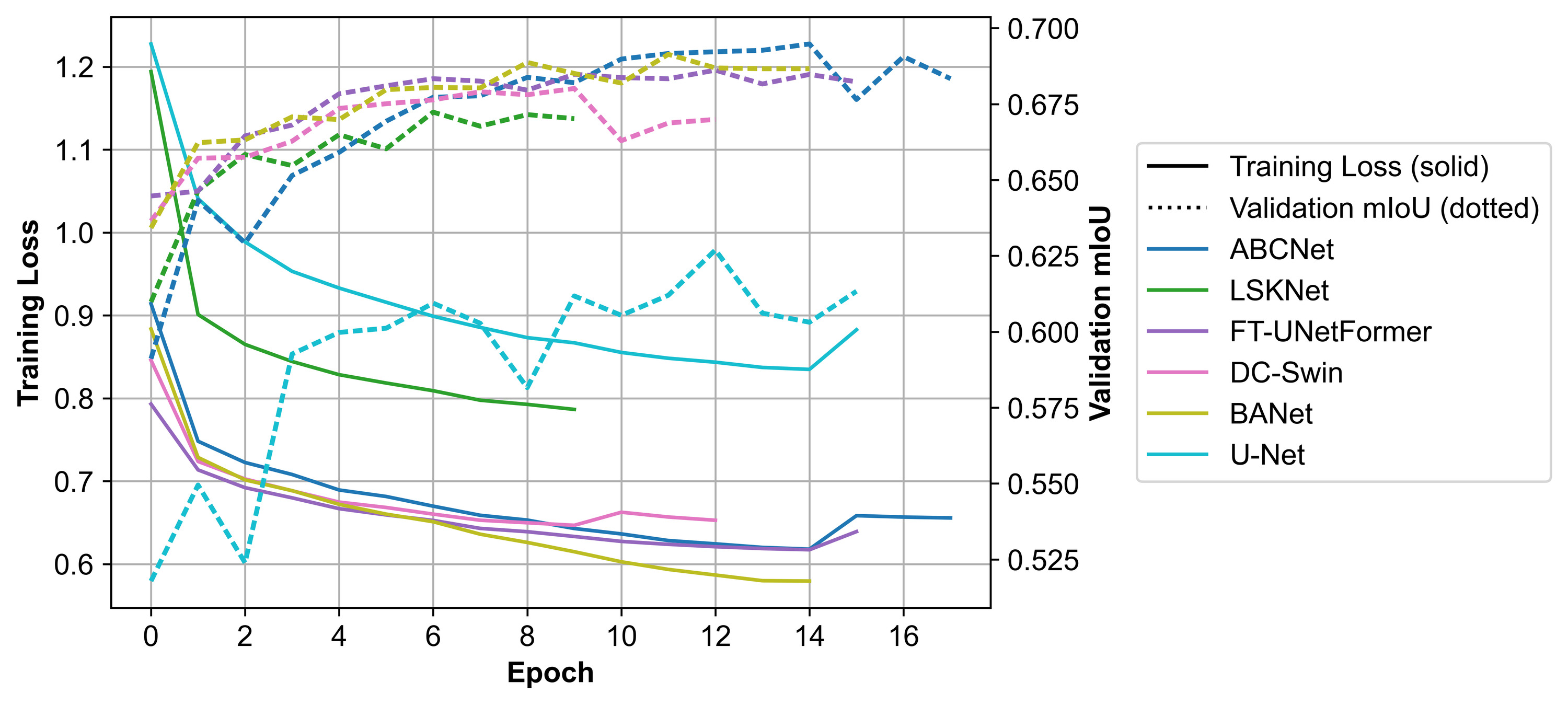}
    \caption{Training loss and validation mean Intersection over Union (mIoU) for the tested models (ABCNet, LSKNet, FT-UNetFormer, DC-Swin, BANet, and U-Net). Solid lines represent training loss, while dashed lines indicate validation mIoU.}
    \label{fig:training}
\end{figure}

\clearpage

\subsection{Detailed Results}

\begin{figure}[!h]
    \centering
    \includegraphics[width=1\textwidth]{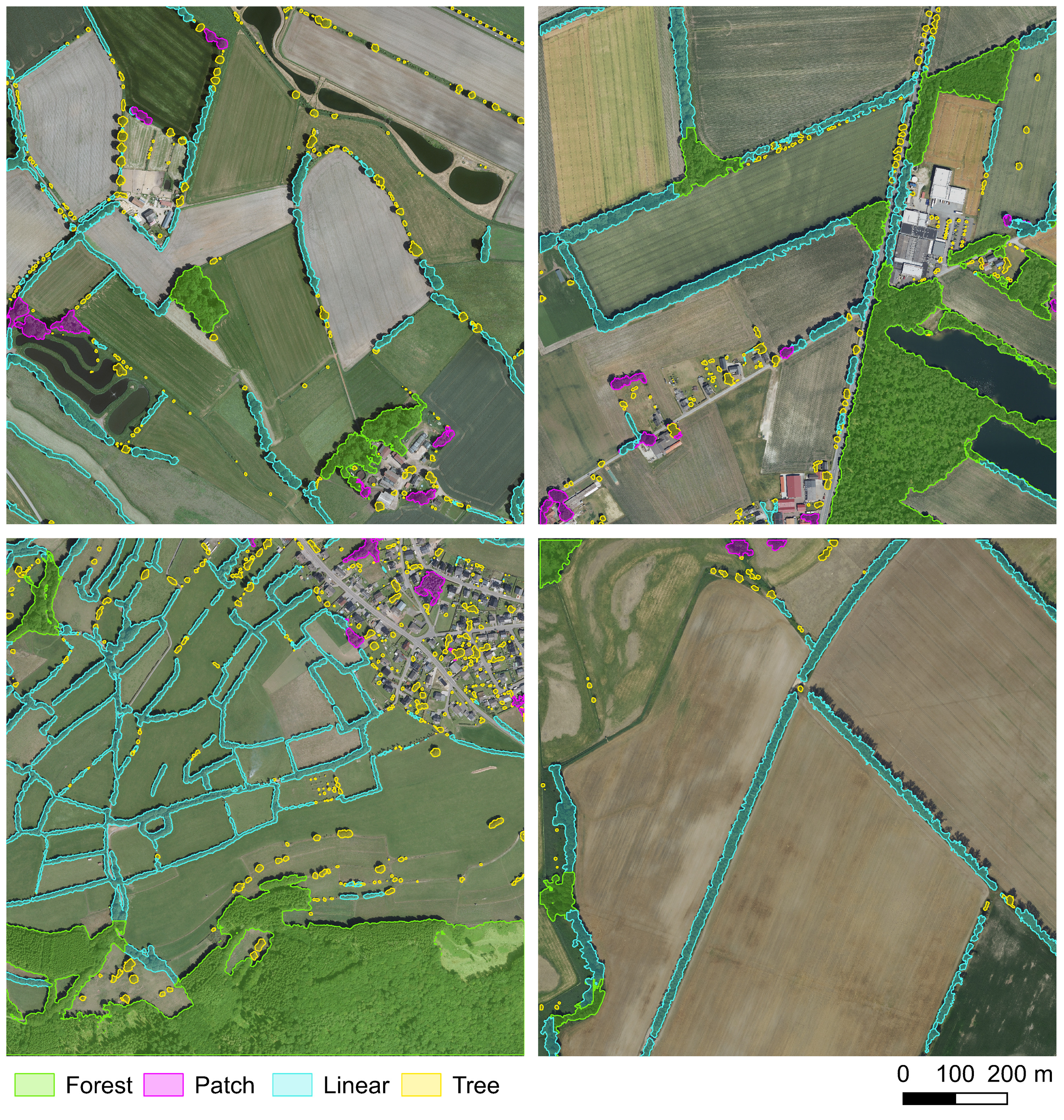}
    \caption{Results of the best-performing model (FT-UNetFormer) in four testing tiles.}
    \label{fig:best_model_detail}
\end{figure}

\clearpage

\subsection{Modality Ablation}
\label{sec:modality_ablation}

To isolate the contribution of each input modality, we trained the same
FT-UNetFormer~\citep{wang_2022} backbone on four variants of the identical
training tiles, differing only in the input channels fed to the patch-embedding
layer. All other settings were kept identical to the training configurations mentioned in \ref{sec:training_setup} across all configurations.

The four configurations were:
\begin{itemize}
  \item \textbf{RGB} (3 bands): aerial orthophoto only; baseline.
  \item \textbf{RGB+NIR} (4 bands): adds the near-infrared channel from the
        same orthophoto.
  \item \textbf{RGB+nDSM} (4 bands): adds the normalised digital surface model
        derived from  stereo image matching.
  \item \textbf{RGB+NIR+nDSM} (5 bands): the full multimodal stack.
\end{itemize}

For variants with more than three input channels, the pretrained Swin-base
patch-embedding convolution was reused for the RGB channels, and the additional
filters were initialised by copying the red-channel weights, providing a
sensible starting point without disrupting the pretrained representation. Each
model was evaluated using the best validation checkpoint (monitored by
class-mean $F_1$) on the held-out test tiles.

Table~\ref{tab:modality_ablation} reports the mean test metrics for the four
configurations. Adding the NIR band changed accuracy only marginally
(+0.4\% mIoU), whereas adding the nDSM produced the largest improvement
(+4.4\% mIoU). Combining both did not exceed RGB+nDSM, indicating that the NIR
band adds little once height information is available.

\begin{table}[h]
\centering
\caption{Mean test metrics of FT-UNetFormer trained with additional input modalities. NIR denotes the additional aerial near-infrared band and nDSM the normalised digital surface model. Values in parentheses give the relative change from the RGB baseline.}
\label{tab:modality_ablation}
\begin{tabular}{lcccc}
\hline
\textbf{Input} & \textbf{mIoU} & \textbf{mF1} & \textbf{mPrecision} & \textbf{mRecall} \\ \hline
RGB            & 0.732 & 0.838 & 0.840 & 0.833 \\
RGB+NIR        & 0.735 & 0.841 & 0.838 & 0.839 \\
RGB+nDSM       & 0.764 & 0.863 & 0.857 & 0.866 \\
RGB+NIR+nDSM   & 0.754 & 0.856 & 0.862 & 0.847 \\ \hline
\end{tabular}
\end{table}

\subsection{Patch Class Oversampling}
\label{sec:oversampling}

The \textit{Patch} class covers only a small fraction of the labeled pixels, so
uniform tile sampling is dominated by Background and Forest tiles and
provides a weak gradient signal for the rare class. We therefore replace uniform
sampling with a class-aware scheme that favors Patch tiles during training,
leaving validation and testing unchanged.

For each training tile we precompute the normalised class histogram. From the mean per-class frequency we define an
inverse-frequency weight restricted to the oversampled classes. Each tile then receives a sampling probability
so Patch-rich tiles are drawn far more often, while tiles without Patch keep a small but positive weight and are not excluded. We draw tiles per epoch with replacement, matching the epoch size of uniform sampling.

\begin{table}[!ht]
    \centering
    \caption{Per-class metrics applying patch oversampling and the change to the baseline in parenthesis.}
    \resizebox{\linewidth}{!}{%
        \begin{tabular}{lllll}\toprule
        Class    & IoU              & F1               & Precision        & Recall           \\ \midrule
        Forest   & 0.904 ($-0.050$) & 0.949 ($-0.027$) & 0.977 ($+0.006$) & 0.922 ($-0.059$) \\
        Patch    & 0.623 ($+0.028$) & 0.774 ($+0.026$) & 0.727 ($-0.066$) & 0.825 ($+0.127$) \\
        Linear   & 0.779 ($+0.006$) & 0.875 ($+0.004$) & 0.922 ($+0.046$) & 0.830 ($-0.038$) \\
        Tree     & 0.674 ($+0.077$) & 0.816 ($+0.060$) & 0.785 ($+0.035$) & 0.850 ($+0.086$) \\ \midrule
        Mean     & 0.750 ($+0.015$) & 0.856 ($+0.016$) & 0.851 ($+0.005$) & 0.863 ($+0.029$) \\
    \bottomrule
    \end{tabular}
    }
    \label{tab:patch_oversampling}
\end{table}

Table~\ref{tab:patch_oversampling} reports the results with Patch-class oversampling.

\clearpage

\bibliographystyle{elsarticle-harv}
\bibliography{references}

\end{document}